\documentclass{article}
\usepackage{spconf,amsmath,graphicx}
\usepackage[ruled,vlined,linesnumbered]{algorithm2e}  
\usepackage[table,xcdraw]{xcolor}

\title{Sentence Boundary Augmentation for Neural Machine Translation Robustness}
\name{Daniel Li$^*$\thanks{$^*$Work done as an intern at Google Research.}, Te I, Naveen Arivazhagan, Colin Cherry, Dirk Padfield}
\address{Google Research}

\begin{document}
%
\maketitle 
\begin{abstract}
Neural Machine Translation (NMT) models have demonstrated strong state of the art performance on translation tasks where well-formed training and evaluation data are provided, but they remain sensitive to inputs that include errors of various types. Specifically, in the context of long-form speech translation systems, where the input transcripts come from Automatic Speech Recognition (ASR), the NMT models have to handle errors including phoneme substitutions, grammatical structure, and sentence boundaries, all of which pose challenges to NMT robustness. 
Through in-depth error analysis, we show that sentence boundary segmentation has the largest impact on quality, and we develop a simple data augmentation strategy to improve segmentation robustness.
\end{abstract}


%
\section{Introduction}
\label{sec:intro}
With the advance of Automatic Speech Recognition (ASR) and Neural Machine Translation (NMT) systems, speech translation has become increasingly feasible and has received considerable attention. 
However, researchers have encountered many challenging problems within the standard cascaded framework where ASR system outputs are passed into NMT systems.
First, since NMT models are often trained with clean, well-structured text, the disfluency of spoken utterances and the recognition errors from ASR systems are not modeled by the NMT systems. 
Second, people speak differently than they write, which results in changes in both sentence structure and meaning. 
Third, automatically predicting sentence boundaries is challenging~\cite{Makhija2019, Nguyen2019Fast, wang-etal-2019-online}. 
Taken as a whole, poorly segmented sentences with incorrect word recognition leads to poor translations. 
These problems pose unique challenges for ASR NMT robustness that are not readily addressed by current methods.

Current approaches to robust NMT with noisy inputs typically focus on improving word transcription through data augmentation techniques. Such methods include disfluency removal~\cite{wang2010} where redundant and unnecessary words are removed before translating the transcript, domain adaptation~\cite{he2011robust} where NMT models are augmented with in-domain training data, and synthetic noise~\cite{cheng-etal-2018-towards,Gangi2019RobustNM}, where random edits are made to training data. 


In our experimentation, we found that ASR system punctuation is often imperfect. It may omit or insert sentence-final punctuation, resulting in sentences that are erroneously compounded or fragmented. 
While this is corroborated by similar works~\cite{Gangi2019RobustNM, Zhang2020DynamicSB}, which note that degradation of translation is caused by poor system sentence boundary prediction, they do not specifically evaluate, quantify, and address this issue.
To tackle the sentence boundary problem, we propose a simple scheme to augment NMT training data, which yields +1 BLEU point on average.
This procedure is agnostic to ASR systems and can be applied to any NMT model training easily.


\section{Experimental Setup}
\label{sec:setup}

\subsection{Data}
We experiment on the IWSLT English to German (EnDe) speech translation scenario.
Training data includes the 4.6M sentence pairs of the WMT 2018 EnDe corpus \cite{bojar-etal-2018-findings} and the IWSLT 2018 EnDe training data, including both the official training set and the leftover TED talks not included in any other test set (0.25M sentence pairs combined). 
For the IWSLT data, we scrape the ground truth transcripts and translations of the above datasets from www.ted.com 
directly, because we found that the official IWSLT datasets omit transcriptions for many sentences, and our focus on segmentation errors requires access to the full long-form spoken document. 
We mix the training data at the mini-batch level with 90\% WMT and 10\% IWSLT, as determined during development. 
Section \ref{sec:pseudocode} provides our data augmentation and mixing details.

We evaluate our models on past IWSLT spoken language translation test sets. We use IWSLT tst2014~\cite{cettolo2014report} as dev set, and test on both IWSLT-tst2015~\cite{Cettolo2015_iwslt} and IWSLT tst2018 \cite{Niehues2018_iwslt}, although our tst2018 includes only the TED talks, not the ``lecture'' files. 
Punctuated ASR transcriptions are obtained from the publicly available Speech-to-Text Google API.\footnote{\texttt{http://cloud.google.com/speech-to-text}}
This achieves a WER of 5.5\% on tst2015 and 6.4\% on tst2018, ignoring case and punctuation.
We run a sentence breaker on the punctuated source to determine the segments translated by NMT.
Since these segments need not match the reference sentence boundaries, especially when punctuation is derived automatically on ASR output, we use a Levenshtein alignment to project the reference boundaries onto our output before evaluating quality with case-sensitive BLEU~\cite{matusov2005evaluating}.

Models are trained in two scenarios: 1) stripped (Train$_s$), where all source sentences are stripped of punctuation, capitalization and symbols, and 2) punctuated (Train$_p$), where source sentences are untouched. Target German translations are untouched in both scenarios, retaining case and punctuation. 
For the evaluation data, source punctuation and case are always preserved or stripped to match the model's training data.
NMT is evaluated with input consisting of either ground truth transcripts and sentence segmentations ($\textit{Gold}$), or ASR transcripts and segmentations ($\textit{System}$).



\subsection{Baseline}
\label{ssec:puncbase}
For all our experiments, we use a Transformer model~\cite{VaswaniSPUJGKP17} with a model dimension of 1024, hidden size of 8192, 16 heads for multihead attention, and 6 layers in the encoder and decoder. The models are regularized using a dropout of 0.3 and label smoothing of 0.1 \cite{SzegedyVISW15}.

To isolate punctuation-related effects from segmentation effects, we first train two NMT models, Train$_s$ and Train$_p$, with results given in Table \ref{tab:baselines}. 
When evaluated on \textit{System} transcripts, we observe a $+1.5$ BLEU gain of Train$_s$ over Train$_p$ by dropping punctuation in training and evaluation. 
This improvement is in line with prior works (training and evaluating translation on stripped data), so we use the $25.8$ BLEU score from the table as our (higher) baseline reference. 
However, we also note that by discarding the punctuation in Gold transcripts, the model's highest attainable performance suffers a $0.5$ BLEU drop.


\begin{table}[t]
\centering
\begin{tabular}{lrl}
\hline \textbf{Training Data} & \textbf{Evaluation Data} & \textbf{BLEU} \\ \hline
Train$_p$ & $\textit{System}$ & $24.3$ \\
Train$_s$ & $\textit{System}$ & $25.8$\\ 
Train$_p$ & $\textit{Gold}$ & $33.5$ \\
Train$_s$ & $\textit{Gold}$ & $33.0$ \\
\hline
\end{tabular}
\vspace{-0.1cm}
\caption{\label{tab:baselines} Baselines on tst2015 dataset. Tran$_s$ is derived from Train$_p$ by stripping punctuation and capitalization from the source. \textit{Gold} and \textit{System} differ in using human generated versus an automatically generated transcript and segmentation for the input. The input is conditionally stripped or lower-cased to match the training data.}
\end{table}


\section{Error Analysis}
\label{sec:erroranalysis}
To motivate our approach of focusing on segmentation errors, we conducted a series of experiments on the tst2015 dataset to explore the impact of noisy segmentation boundaries on machine translation. 
We demonstrate that translation degradation due to segmentation errors is more than double that due to transcript recognition errors. 

\begin{table}[t]
\centering
\begin{tabular}{ll}
\hline \textbf{Data Type} & \textbf{Example Sentence} \\ \hline
$\textit{Gold}$ & the \textbf{weather} today was warm $\mathbf{\langle s \rangle}$ \\
$\textit{System}$ & the \textbf{whether}  $\mathbf{\langle s \rangle}$ today was warm  $\mathbf{\langle s \rangle}$\\  \hline
$\textit{Recognition}$ & the \textbf{whether} today was warm  $\mathbf{\langle s \rangle}$ \\
$\textit{Segmentation}$ & the \textbf{weather}  $\mathbf{\langle s \rangle}$ today was warm  $\mathbf{\langle s \rangle}$ \\
\hline
\end{tabular}
\vspace{-0.1cm}
\caption{\label{tab:projection_example} An example of our two inputs (Gold and System) and how we can create two more
with isolated error categories by projecting sentence boundaries $\mathbf{\langle s \rangle}$ from one to the other.}
\end{table}
\subsection{Sentence Boundary Projection}
\label{ssec:projectpunct}
In order to isolate sentence boundary errors from recognition errors, we project sentence boundaries from \textit{System} transcripts to \textit{Gold} transcripts and vice versa. 
We begin with the novel insight that we can use the same Levenshtein-driven boundary projection algorithm that we are already comfortable with using for evaluation~\cite{matusov2005evaluating} to enable our error analysis.
Specifically, we can perform a token-level Levenshtein alignment ignoring segment boundaries, and then transfer the boundaries from sentence $A$ to another sentence $B$ according to the alignments of the tokens surrounding the boundary symbols in $A$.

If we use this technology to project Gold boundaries onto System, 
we get (only) \textit{Recognition Errors}, indicating a model's performance on perfect sentence boundaries and noisy system transcripts. If we project System boundaries onto Gold, we get (only) \textit{Segmentation Errors}, indicating a model's performance on noisy system sentence boundaries and perfect transcripts. Examples for both projections are provided in the bottom half of Table~\ref{tab:projection_example}. We evaluate the punctuation-stripped Train$_s$ NMT on \textit{Recognition Errors} and \textit{Segmentation Errors} as a proxy to quantify the effects of different types of noise on translation accuracy and to identify areas for improvement.


\subsection{Transcription Sentence Boundary Error Analysis}
\label{ssec:mixedeffects}

\begin{table}[h]
\centering
\begin{tabular}{lrl}
\hline
\textbf{Evaluation Data} & \textbf{BLEU} \\
\hline
\textit{Gold} (No Errors) & $33.0$\\ 
\textit{Recognition Errors} & $31.0$\\ 
\textit{Segmentation Errors} & $28.7$\\ 
\textit{System} (Both Errors) & $25.8$\\ 
\hline
\end{tabular}
\vspace{0.2cm}
\caption{\label{tab:headroom} Train$_s$ evaluation on tst2015 for different combinations of transcript type and segmentation.}
\end{table}

Table \ref{tab:headroom} gives the BLEU scores on the evaluation data created in Section \ref{ssec:projectpunct} using the same Train$_s$ model from Table \ref{tab:baselines}.
When compared to the \textit{Gold} transcript ($33.0$ BLEU), we observe the highest degradation occurs on \textit{Segmentation Errors}, by $-4.3$ BLEU points, while \textit{Recognition Errors} degrade translation by only $-2.0$ BLEU points. 
These observations motivate our approach to focus on segmentation errors, especially considering how, to the best of our knowledge, all other current methods for achieving NMT robustness focus on recognition errors as opposed to segmentation errors.



\section{Data Augmentation for Segmentation Robustness}
\label{sec:pseudocode}
It makes sense that sentence boundary errors could affect translation quality: incorrect boundaries could separate words from context that is critical for their correct translation.
However, with such a large effect, we were concerned that the NMT system was unnecessarily sensitive to the unnatural boundaries  from automatic punctuation; that is, even with sufficient context, the system could be erring simply because the sentence boundary appeared in an unexpected place.
Therefore, we propose a data augmentation procedure to expose the model to bad segmentations during training.

Pseudocode for our procedure is given in Algorithm \ref{sec:alg1}, which represents sentences as sequences of tokens. Adjacent source sentences $S$ are concatenated, as are their corresponding target sentences $T$.
Next, the start and end of the concatenated source and target sentences are proportionally\footnote{Due to linguistic mappings between translations, the concatenated source and target sentences differ in length, so the truncation start and end indices are proportionally and independently obtained ($\texttt{lines 5,6,7,8}$).} truncated to imitate a random start or break. 
This is governed by the hyperparameter $p=0.3$. By design, the truncation keeps most of the first sentence ($\texttt{lines 7,8}$ for $S_1, T_1$: at most 30\% of the start of the first sentence is discarded) while discarding the bulk of the second sentence ($\texttt{lines 7,8}$ for $S_2, T_2$: at most 30\% of the start of the second sentence is retained). This removes context from the first sentence $S_1$ and adds context from the second sentence $S_2$ and combines them into a single training example.
This data augmentation strategy is similar to the prefix training methods proposed in \cite{Niehues2018_low, ma-etal-2019-stacl}, but we extend it to a sentence-level prefix and suffix subsampling across sentence boundaries.



\begin{algorithm}[h]
\label{sec:alg1}
\caption{Data Augmentation}
\KwIn{List of source $S$, target $T$ sentences as tuples $L=\{(S_i, T_i)\}_{i=0}^{n}$,  hyper-parameters $p=0.3$ for sentence length truncation.}
$L_{out} \gets $ list()\;
\For{$i=0,2,4,..,n-2,n$}{
 $p \sim  \text{Uniform}(0, p)$\;
 $(S_1, T_1), (S_2, T_2) = L[i], L[i+1]$\;
 $p_1^{S_1}, p_1^{T_1}  = \lceil p \times $len$(S_1)\rceil, \lceil p \times $len$(T_1)\rceil$\;
 $p_2^{S_2}, p_2^{T_2}  = \lceil p \times $len$(S_2)\rceil,  \lceil p \times $len$(T_2)\rceil$\;
 $S_{out} = $ concat$([S_1[p_1^{S_1}:], S_2[:p_2^{S_2}]])$\;
 $T_{out} = $ concat$([T_1[p_1^{T_1}:], T_2[:p_2^{T_2}]])$\;
 $L_{out}$.append$([S_{out}, T_{out}])$\;
 }
\Return $L_{out}$
\end{algorithm}


We augment WMT and IWSLT separately, creating two punctuation-stripped corpora with 80\% original full-sentence data and 20\% newly created augmented data. As in Section~\ref{sec:setup}, for each batch we sample 90\% data from WMT (with augmentation) and 10\% IWSLT (with augmentation).

\section{Results}
\label{sec:mainresults}

We evaluate  Train$_s$ with and without data augmentation in a variety of settings. 
We demonstrate the robustness of our approach to segmentations derived from automatic punctuation, but also to those from speaker pauses and fixed-length strategies. Training regimen, hyper-parameter selection, and architecture are all identical across all experiments. 



\subsection{Augmentation Evaluation}
\label{sec:tedtalk}
To evaluate the augmentation's effectiveness, we use the ASR \textit{System} transcript, evaluating both tst2015 and tst2018.
As shown in Table \ref{tab:tedtalks}, our augmentation scheme improves the translation accuracy by $+1.2$ BLEU and $+0.7$ BLEU points, respectively. 

\begin{table}[ht!]
\centering
\begin{tabular}{lrl}
\hline \textbf{Training Data} & \textbf{tst2015} & \textbf{tst2018} \\ \hline
Train$_s$ & $25.8$ & $21.3$ \\
Train$_s$ + Augmentation & $\mathbf{27.0}$ & $\mathbf{22.0}$ \\
\hline
\end{tabular}
\caption{tst2015 and tst2018 \textit{System} transcripts BLEU scores.}\label{tab:tedtalks}\vspace{-0.3cm}
\end{table}

\begin{figure}[!h]
\centering
\includegraphics[scale=0.4]{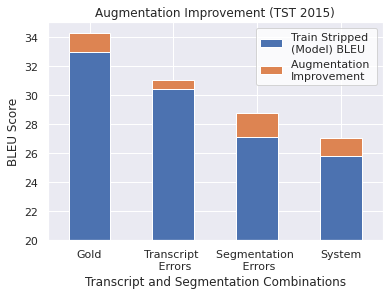}
\caption{Train$_{s}$ + Augmentation differential comparisons across different evaluation transcript scenarios for tst2015.}\label{fig:diffs}
\end{figure}

Figure \ref{fig:diffs} shows the impact of Train$_s$ + Augmentation on robustness to segmentation errors as opposed to recognition errors using the previously created tst2015 \textit{Recognition Error} and \textit{Segmentation Error} evaluation transcripts from Section~\ref{ssec:mixedeffects}.
We observe a clear increase in performance on the \textit{Segmentation Error} transcript compared to other scenarios, suggesting the model increases robustness towards ASR system sentence boundary errors. Evaluating the \textit{Gold} transcripts, we note that Train$_s$ + Augmentation ($34.2$ BLEU) recovers the non-augmented model's degradation from punctuation stripping (Table \ref{tab:baselines}) and further outperforms Train$_p$ by $+0.3$ BLEU, increasing the NMT model's best attainable translation performance on Gold data.

\begin{figure}[h]
\centering
\includegraphics[scale=0.4]{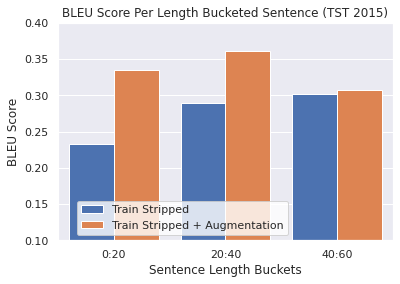}
\vspace{-0.2cm}
\caption{Average BLEU score of sentences in specific token length bucket.}\label{fig:sentences}\vspace{-0.2cm} 
\end{figure}

Finally, we inspect the impact of augmentation on sentences of different lengths. Figure \ref{fig:sentences} depicts a side-by-side comparison of a model's aggregated BLEU score for sentences with lengths contained in buckets $\{0:20\}, \{20:40\}, \{40:60\}$.
This shows that translated sentences in buckets $\{0:20\}$ and $\{20:40\}$ benefit most from augmentation. 

\subsection{Robustness to Alternative Segmentation Strategies}

We study how well our model performs with some simple segmentation heuristics instead of using a system punctuator. Our findings demonstrate that the proposed augmentation strategy is robust under a wide range of segmentation approaches.

\subsubsection{Speaker Pause}
\label{sec:asrapplication}
Here we utilize a feature of the ASR system that 
splits a transcription whenever it detects a long enough speaker pause (1 sec). 
Due to prose variance, the split can occur mid-sentence if a speaker pauses before completing a sentence. 
If a segment is too long, we further split it so the final segments are no longer than (50, 70, 90) tokens. 
We then use Train$_s$ to translate and evaluate these simulated segments. 
Under this schema, the NMT system ignores the sentence boundaries information provided by automatic punctuation.

\begin{figure}[h!]
\centering
\includegraphics[scale=0.4]{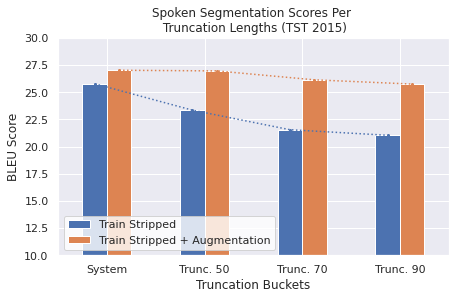}
\vspace{-0.2cm}
\caption{\label{fig:segments} Models evaluated on tst2015 \textit{System} transcripts or by speaker pauses and a fixed-length split.}\vspace{-0.3cm}
\end{figure}

Surprisingly, even with this na\"{i}ve pause-based segmentation strategy, our augmented model can achieve the same performance as when sentence boundaries are given by a dedicated punctuation model.
This is shown in Figure \ref{fig:segments} where \textit{System} and Trunc 50 have the same performance when using Train$_s$ + Augmentation. Furthermore, the augmented NMT model remains relatively robust given even longer truncation thresholds compared to the non-augmented model.

\subsubsection{Length Segmentation}
\label{sec:robustness}
Next we segment the text by simply counting tokens. \textit{System} transcripts are greedily split into segments with a fixed number of tokens, translated, and evaluated. Segment lengths vary from 10 to 100.
We observe Train$_s$ + Augmentation maintains better performance when compared to Train$_s$, reaching a peak just shy of $1$ BLEU point from its absolute best \textit{System} transcript performance, further indicative of our data augmentation's robustness to ASR segmentation errors.

\begin{figure}[h!]
\centering
\includegraphics[scale=0.38]{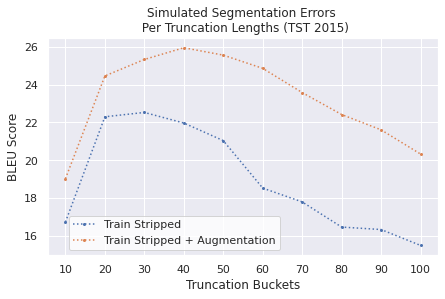}
\vspace{-0.2cm}
\caption{\textit{System} transcripts generated with incorrect segmentations. All models suffer a marked drop at 10 tokens.}
\end{figure}

\section{Conclusion}
\label{sec:conclusion}
We propose that sentence boundary errors are a neglected area of study for NMT robustness, especially in the context of speech translation. 
We quantitatively demonstrate that poor sentence segmentation degrades performance almost twice as much as transcript level-errors. 
To address this, we developed a simple method for data augmentation with immediate gains that can serve as a baseline for future work in segmentation NMT robustness. 

\vfill\pagebreak

\label{sec:refs}
\bibliographystyle{IEEEbib}
\bibliography{refs}

\end{document}